\documentclass{article}
\usepackage{spconf,amsmath,graphicx}
\usepackage{amssymb}
\usepackage{booktabs}
\pdfoutput=1

\title{AuthFormer: Adaptive Multimodal biometric authentication transformer for middle-aged and elderly people}
\name{Rui Yang, Qiuyu Zhang$^{*}$, Lingtao Meng\thanks{This work was supported in part by the National Natural Science Foundation of China under Grant 61862041.}}
\address{School of Computer and Communication, Lanzhou University of Technology, Lanzhou, 730050, China \\
      \{rykeryang, zhangqy, menglt\}@lut.edu.cn 
}

\begin{document}
%
\maketitle
\begin{abstract}
Multimodal biometric authentication methods address the shortcomings of unimodal biometric authentication technologies regarding security, robustness, and user adaptability. However, most existing multimodal biometric authentication methods rely on fixed biometric modality combinations and fixed numbers of biometric modalities, which greatly restrict the flexibility and adaptability of such systems in real-world applications. To tackle these challenges, we proposes an adaptive multimodal biometric authentication model (AuthFormer) specifically designed for elderly populations. AuthFormer is trained on the LUTBIO multimodal biometric database, which includes biometric data from elderly individuals. By integrating a cross-attention mechanism and a Gated Residual Network (GRN), the model significantly enhances its adaptability to physiological feature variations specific to elderly users. Experimental results demonstrate that the AuthFormer model achieves an accuracy of 99.73\%. Moreover, the AuthFormer model’s encoder requires only two layers to achieve optimal performance, significantly reducing model complexity compared to traditional Transformer-based authentication models.
\end{abstract}
\section{Introduction}
\label{sec:intro}

In the field of multimodal biometric authentication, the effective integration of multiple biometric features to enhance the accuracy and robustness of identity verification has emerged as a key issue in advancing next-generation security technologies ~\cite{kumar2022comprehensive,mamdouh2021authentication}. Multimodal biometric authentication incorporates various biometric modalities, such as fingerprint, face, and iris, to improve the reliability and security of identity verification by leveraging information from different modalities. Substantial progress has been achieved in the research and application of multimodal biometric recognition/authentication in recent years~\cite{ante2022bibliometric}.

The fusion of biometric modalities in multimodal biometric authentication typically takes place at the sensor, feature, score, and decision levels~\cite{sharma2023survey}. Existing research primarily focuses on selecting appropriate biometric modalities for fusion-based authentication at different stages~\cite{li2023imf, talreja2020deep}. However, these approaches face three significant challenges: 1)Inadequate exploitation of biometric modality information and the relationships among different modalities, leading to difficulties in achieving complementary and enhanced integration of biometric features during the fusion process~\cite{choudhury2021adaptive, al2023keystroke}; 2) Limited research addressing multimodal biometric authentication for specific populations, such as elderly individuals~\cite{chun2024healthspan}; 3) A shortage of studies on multimodal biometric authentication systems that allow users to dynamically select suitable biometric modalities and combinations. Current studies evaluate the benefits of multimodal biometric authentication models across different fusion stages, typically focusing either on which biometric modalities are integrated at which stages~\cite{lien2023challenges} or on adaptive biometric authentication~\cite{arias2019survey}. However, none simultaneously explore adaptive multimodal biometric authentication specifically tailored for elderly populations.

Therefore, we have developed a multimodal biometric database, LUTBIO~\cite{yang2024lutbio}, which includes biometric data from elderly individuals, and proposed an efficient adaptive multimodal biometric authentication method based on this database. Our primary contributions are as follows: 

\begin{itemize}
    \item Design a general multimodal biometric authentication framework model (AuthFormer) that accommodates various combinations and quantities of biometric modalities while incorporating a novel embedding method for encoding sequential biometric data; 
    \item Propose a multimodal fusion method incorporating GRN and a cross-attention mechanism to fully enable complementary integration of various biometric features, along with an adaptive module that dynamically delivers identity authentication services based on the biometric modalities provided by the user.
\end{itemize}

\begin{figure*}[!h]
\centerline{\includegraphics[width=7in]{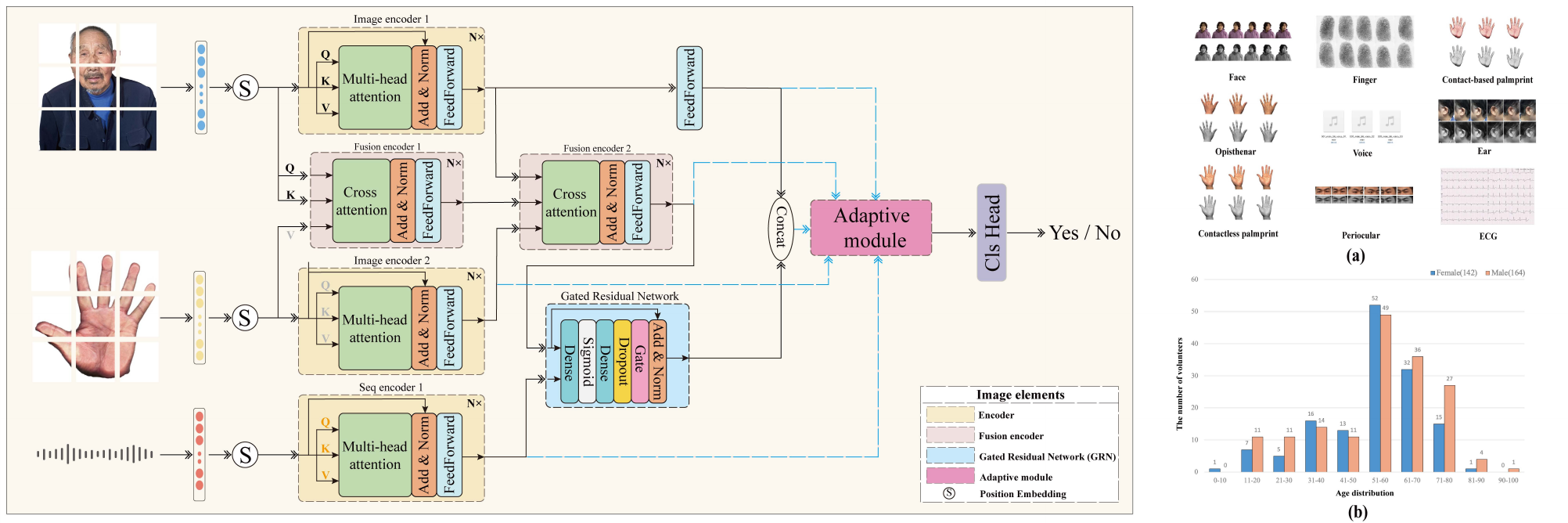}}
\caption{The left side of the figure shows the framework of AuthFormer, while the right side displays the biometric modalities (a) and subject age distribution (b) included in the LUTBIO multimodal biometric database.}
 \label{fig1}
\end{figure*}

\section{Proposed method}
\label{sec:format}

\subsection{Embedding method}
Given a multimodal biometric dataset, the system incorporates three image biometric modalities and one sequential biometric modality, aiming to design an embedding method for the adaptive multimodal biometric authentication model, AuthFormer, that effectively leverages information from each biometric modality. We adopt the Vision Transformer~\cite{dosovitskiy2020image} embedding method for the image modalities and propose a novel embedding method for sequential biometric features. For the sequential biometric features $x(t) \in \mathbb{R}^{T}$ (where $T$ represents the sequence feature length), the process of using a learnable embedding to obtain the embedded vector $e = [e_1, e_2, ..., e_N]$ is as follows(where $N$ denotes the sequence length of the embedded vectors, with each $e_i$ having a dimension of $d$):
\begin{equation}
    E(x(t)) = e \in \mathbb{R}^D
\end{equation}
where $D$ denotes the dimension of the embedded vector $e$, and $E$ represents the learnable embedding function. 

Then, the process of extracting features from the embedded vector $e$ using a Temporal Convolutional Network(TCN) is as follows:
\begin{equation}
     F(e) = f \in \mathbb{R}^D
\end{equation}
where $F$ represents the function of the TCN model for feature extraction, and $f$ is the feature vector.

Finally, positional encoding is added to each feature to obtain the final sequential feature input, the process is as follows:
\begin{equation}
     S_i =  \left\{ f_i + p_i \mid i = 1, 2, ..., N \right\}
\end{equation}
where $f$ is the $i$-th feature, $P$ is the positional encoding of the $i$-th feature, and $S$ is the result of adding the positional encoding to the $i$-th feature.

\subsection{Attention mechanism}
As shows in Fig.~\ref{fig1}, AuthFormer employs two types of attention modules: a self-attention module and a cross-attention module. The self-attention mechanism is used to independently process image-based biometric modalities $Z^{face} = [Z_1^{face}, Z_2^{face}, ..., Z_N^{face}] \in \mathbb{R}^{N \times D}$ and sequential biometric modalities $S^{voice} = [S_1^{voice}, S_2^{voice}, ..., S_N^{voice}] \in \mathbb{R}^{N \times D}$, where $N$ represents the number of patches, and $D$ denotes the embedding dimension. 

The cross-attention module is responsible for fusing multiple image-based biometric modalities, and AuthFormer employs this module twice. The first cross-attention module handles the initial fusion of the two image-based biometric modalities. The intermediate information generated from this fusion is then processed by the second cross-attention module fully utilizing the information from both image-based biometric modalities to generate fused information. The process is as follows:

\begin{equation}
\begin{aligned}
B^{'} = CrossMSA(Q^{face}, K^{face}, V^{finger})\\+ (Q^{face}, K^{face}, V^{finger})
\end{aligned}
\end{equation}

\begin{equation}
    B^{stage1} = MLP(LN(B^{'}))\\+ (B^{'})
\end{equation}

\begin{equation}
\begin{aligned}
B^{''}  = CrossMSA(B^{face}, B^{stage1}, B^{finger}) \\+ (B^{face}, B^{stage1}, B^{finger})
\end{aligned}
\end{equation}

\begin{equation}
    B^{fusion} = MLP(LN(B^{''}))+B^{''}
\end{equation}

\subsection{Fusion method based on Gating Mechanisms}

To enable biometric features to complement and reinforce each other during the fusion process, AuthFormer utilizes Gated Residual Network (GRN)~\cite{huang2020automatic} to fuse image modality information with sequential modality information. As shown in Fig.~\ref{fig1}, the GRN module incorporates a component gating layer based on the Gated Linear Unit (GLU)~\cite{liu2023time} to introduce flexibility. This design aims to strengthen the model's focus on important features while diminishing attention to less significant ones. The process of GRN module is as follows:

\begin{equation}
    GRN_w(B^{fusion}, S^{voice}) = LN(B^{fusion} + GLU_w(m_1))
\end{equation}
\begin{equation}
    m_1 = W_{1,w} \cdot m_2 + b_{1,w}
\end{equation}
\begin{equation}
    m_2 = Sigmoid(W_{2,w} \cdot B^{fusion} + W_{3,w}\cdot S^{voice} + b_{2,w})
\end{equation}
\begin{equation}
    GLU_{w}(n) = sigmoid(W_{4,w} \cdot n \\+ b_{4,w}) \odot (W_{5,w} \cdot n + b_{5,w})
\end{equation}
where $sigmoid$ is the activation function, $m_1 \in \mathbb{R}^{d_{model}}$ and $m_2 \in \mathbb{R}^{d_{model}}$ are the intermediate layers, $LN$denotes the layer normalization function, $w$ serves as the weight sharing index, and $W_{( \cdot )} \in \mathbb{R}^{d_{model} \times d_{model}}$ and $b_{ (\cdot) }  \in \mathbb{R}^{d_{model}}$ represent the weights and bias, respectively.

\subsection{Adaptive module}

The adaptive module is a critical component of the AuthFormer model, enabling the system to dynamically adjust the authentication process according to the user’s specific needs and environmental conditions. AuthFormer incorporates two adaptive capabilities: adaptive biometric modality quantity and adaptive biometric modality combination. The adaptability of biometric modality quantity allows users to provide up to three different biometric modalities for authentication based on their preferences and environmental factors, such as fingerprint wear. The adaptability of biometric modality combination allows AuthFormer to bypass the constraints of fixed biometric modality combinations, enabling users to freely choose any combination of modalities for authentication. As indicated by the blue dashed line in Fig.~\ref{fig1}, the adaptive module can process single image biometric modality information, sequential biometric modality information, fused image modality information, and information from the fusion of three biometric modalities, ultimately generating the final classification information based on the user-provided biometric modalities. 
\section{Experiment}
\label{sec:pagestyle}
\subsection{Datasets and Implementations}

To evaluate the effectiveness of the AuthFormer,  extensive experiments are conducted on two available multimodal biometric databases. 1)LUTBIO~\cite{yang2024lutbio}: As shown on the right side Fig.~\ref{fig1}, the LUTBIO database is a self-constructed multimodal biometric database focused on the elderly population, encompassing nine types of biometric data: voice, face, fingerprint, contact-based palmprint, electrocardiogram (ECG), opisthenar, ear, contactless palmprint, and periocular. The LUTBIO database consists of 306 subjects. 2)XJTU~\cite{XJTU}: It contains 102 volunteers, including face, voice, contact-based palmprint, and fingerprint as the four biometric modalities.

We compare AuthFormer with four comparison methods, including a score-level fusion model~\cite{Android}, a multimodal score-level fusion method utilizing voice quality assessment~\cite{mobile}, an adaptive fusion method~\cite{zhang2017low}, and a privacy-preserving authentication method~\cite{Privacy}. We use True Acceptance Rate (TAR), False Rejection Rate (FRR), False Acceptance Rate (FAR), Equal Error Rate (ERR), accuracy (acc), macro F1 score, and macro recall as the evaluation metric to validate the performance of the model.

\subsection{Experimental result and Analysis}
\subsubsection{Comparative Experiment}

As shown in Table~\ref{tab1}, the AuthFormer achieves the highest scores in accuracy (0.9973), macro recall (0.9972), and macro F1 score (0.9977), surpassing all unimodal authentication models by at least 2 percentage points. This comparison further emphasizes the effectiveness of the AuthFormer model in enhancing recognition accuracy compared to unimodal authentication models.

\begin{table}[h]
\centering
\caption{The performance of various models in multimodal and unimodal biometric authentication tasks.}
\resizebox{\columnwidth}{!}{
\begin{tabular}{lccc}
\toprule
\textbf{Model} & \textbf{Accuracy} & \textbf{Macro Recall} & \textbf{Macro F1} \\
\midrule
AuthFormer (Face \& Finger \& Voice) & \textbf{0.9973} & \textbf{0.9972} & \textbf{0.9977} \\
ViT[17] (Face) & 0.9724 & 0.9752 & 0.9733 \\
ViT[17] (Finger) & 0.9701 & 0.9721 & 0.9709 \\
GRU (Voice) & 0.5915 & 0.5915 & 0.5859 \\
MobileNet (Face) & 0.9235 & 0.9233 & 0.9235 \\
MobileNet (Finger) & 0.9327 & 0.9257 & 0.9317 \\
\bottomrule
\end{tabular}
}
\label{tab1}
\end{table}

\subsubsection{Ablation study}
To verify the effectiveness of each module, we conduct an ablation study on the LUTBIO database, using varying quantities of biometric modality data. As shown clearly in Table~\ref{tab2}, the accuracy of the AuthFormer model based on fingerprint and voice modalities and the model based on face and voice modalities are 0.9431 and 0.9653, respectively, demonstrating the GRN network’s effectiveness in fusing image and sequential information. Moreover, the macro F1 score of the AuthFormer model based on face and fingerprint modalities is 0.9888, which surpasses that of the unimodal authentication model. This outcome confirms that the cross-attention module effectively fuses multiple image modalities, thereby enhancing the model’s recognition capability.

\begin{table}[h]
\centering
\caption{The performance of AuthFormer uses different quantities and combinations of biometric modalities.}
\resizebox{\columnwidth}{!}{
\begin{tabular}{lccc}
\toprule
\textbf{Combination} & \textbf{Accuracy} & \textbf{Macro F1} & \textbf{Macro Recall} \\
\midrule
Palmprint \& Finger \& Voice & 0.9665 & 0.9652 & 0.9661 \\
Face \& Palmprint \& Voice & 0.9781 & 0.9790 & 0.9791 \\
Finger \& Face \& Voice & \textbf{0.9973} & \textbf{0.9977} & \textbf{0.9972} \\
Face \& Palmprint & 0.9793 & 0.9711 & 0.9781 \\
Face \& Voice & 0.9653 & 0.9600 & 0.9671 \\
Finger \& Palmprint & 0.9503 & 0.9512 & 0.9574 \\
Finger \& Voice & 0.9431 & 0.9389 & 0.9417 \\
Finger \& Face & 0.9888 & 0.9813 & 0.9870 \\
Palmprint \& Voice & 0.9323 & 0.9300 & 0.9321 \\
Face & 0.9678 & 0.9718 & 0.9700 \\
Palmprint & 0.9601 & 0.9629 & 0.9598 \\
Finger & 0.9821 & 0.9846 & 0.9827 \\
Voice & 0.9201 & 0.9198 & 0.9233 \\
\bottomrule
\end{tabular}
}
\label{tab2}
\end{table}

The performance of the AuthFormer model improves significantly as the number of biometric modalities increases. When only the face modality is used, the accuracy is 0.9678. When both face and voice modalities are combined, the accuracy reaches 0.9888, marking a 0.021 increase over the face-only model. When three modalities—fingerprint, face, and voice—are utilized, the accuracy peaks at 0.9973, which is 0.0295 higher than when using the face modality alone. These results indicate that the adaptive module in the AuthFormer model can flexibly adjust the fusion strategy based on the number of input modalities, fully leveraging multimodal information to enhance overall recognition performance.

The data in Table~\ref{tab2} reveal that different modality combinations have a significant impact on model performance. The combination of fingerprint, face, and voice modalities achieves the highest accuracy of 0.9973, making it the best-performing combination. The combination of palmprint, fingerprint, and voice also performs well, with an accuracy of 0.9665, slightly lower than that of the fingerprint, face, and voice combination. This suggests that the adaptive module in the AuthFormer model can effectively adjust to different modality combinations.

\subsubsection{Efficiency analysis}

\begin{figure}[htb]
\begin{minipage}[b]{1.0\linewidth}
  \centering
  \centerline{\includegraphics[width=8.5cm]{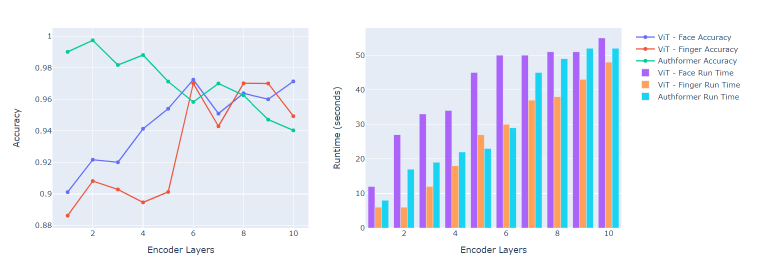}}
\end{minipage}
\caption{Comparison of accuracy and time consumption for different encoder layers.}
\label{fig2}
\end{figure}

To assess the performance of the AuthFormer model in terms of encoder layer count and running efficiency, experiments were conducted using the LUTBIO database, comparing the accuracy and running time of the ViT model and AuthFormer model under different encoder layer settings. The results of the efficiency analysis are shown in Fig.~\ref{fig2}. The accuracy of the ViT model in biometric authentication tasks steadily improves as the number of encoder layers increases, but the training time per epoch also rises significantly. In contrast, the AuthFormer model achieves optimal accuracy with only 2 encoder layers, while its running time remains substantially lower than that of the other models. AuthFormer not only ensures high precision but also demonstrates higher operational efficiency, making it highly promising for practical deployment.

\subsubsection{Generalization Study}

As shown in Table~\ref{tab3}, AuthFormer demonstrates outstanding performance on Four evaluation metrics on the XJTU database~\cite{XJTU}. Specifically, the TAR of the AuthFormer model is 100.00\%, which is 0.39\% higher than the 99.61\% achieved by the Aronowitz model~\cite{mobile}. Additionally, the FRR of the AuthFormer model is 0.00\%, 0.016\% lower than the 0.016\% of the Li et al.~\cite{Privacy}(or) model. Regarding FAR, the AuthFormer model achieves 0.00\%, significantly lower than the 0.661\% recorded by the Li et al.~\cite{Privacy}(and) model, with a difference of 0.661\%. While the works of Zhang et al.~\cite{Android} and Heng et al.~\cite{zhang2017low} achieved performance similar to the AuthFormer model, the AuthFormer model offers support for more biometric modalities and adaptive authentication.

\begin{table}[h]
\centering
\caption{Comparison of TAR, FRR, FAR and ERR for different methods on the XJTU database.}
\resizebox{\columnwidth}{!}{
\begin{tabular}{lcccc}
\toprule
\textbf{Method} & \textbf{TAR} & \textbf{FRR} & \textbf{FAR} & \textbf{ERR} \\
\midrule
Zhang et al.~\cite{Android} & 100.00 & 0.00 & 0.00 & - \\
Aronowitz et al.~\cite{mobile} & 99.61 & 0.39 & 0.20 & - \\
Heng et al.~\cite{zhang2017low} & 100.00 & 0.00 & 0.00 & - \\
Li et al.~\cite{Privacy}(or) & - & 0.016 & 0.054 & 0.035 \\
Li et al.~\cite{Privacy}(and) & - & 0.661 & 0.017 & 0.339 \\
\textbf{AuthFormer (Ours)} & \textbf{100.00} & \textbf{0.00} & \textbf{0.00} & \textbf{0.00} \\
\bottomrule
\end{tabular}
}
\label{tab3}
\end{table}

\section{Conclusion}
\label{sec:typestyle}
This study aims to develop a flexible and adaptive multimodal biometric authentication model, AuthFormer, tailored for elderly populations. By integrating multiple biometric modalities, this model improves system security, adaptability, and robustness, overcoming the limitations of fixed modality combinations found in traditional systems. AuthFormer, utilizing a cross-attention mechanism and GRN, achieves an authentication accuracy of 99.73\% in experiments with face, fingerprint, and voice biometric modalities. Remarkably, AuthFormer reaches optimal performance with just 2 encoder layers, significantly reducing model complexity. Future research explores integrating privacy protection mechanisms.

\vfill\pagebreak

\label{sec:refs}
\bibliographystyle{IEEEbib}
\bibliography{authformer}

\begin{thebibliography}{10}

\bibitem{al2023keystroke}
Jaafer Al-Saraireh and Mohammad~Rasool AlJa'afreh.
\newblock Keystroke and swipe biometrics fusion to enhance smartphones authentication.
\newblock {\em Computers \& Security}, 125:103022, 2023.

\bibitem{ante2022bibliometric}
Lennart Ante, Constantin Fischer, and Elias Strehle.
\newblock A bibliometric review of research on digital identity: Research streams, influential works and future research paths.
\newblock {\em Journal of Manufacturing Systems}, 62:523--538, 2022.

\bibitem{arias2019survey}
Patricia Arias-Cabarcos, Christian Krupitzer, and Christian Becker.
\newblock A survey on adaptive authentication.
\newblock {\em ACM Computing Surveys (CSUR)}, 52(4):1--30, 2019.

\bibitem{mobile}
Hagai Aronowitz, Min Li, Orith Toledo-Ronen, Sivan Harary, Amir Geva, Shay Ben-David, Asaf Rendel, Ron Hoory, Nalini Ratha, Sharath Pankanti, and David Nahamoo.
\newblock Multi-modal biometrics for mobile authentication.
\newblock In {\em IEEE International Joint Conference on Biometrics}, pages 1--8, 2014.

\bibitem{choudhury2021adaptive}
Surabhi~Hom Choudhury, Amioy Kumar, and Shahedul~Haque Laskar.
\newblock Adaptive management of multimodal biometrics—a deep learning and metaheuristic approach.
\newblock {\em Applied Soft Computing}, 106:107344, 2021.

\bibitem{chun2024healthspan}
Elizabeth Chun, Annie Crete, Christopher Neal, Richard Joseph, and Rachele Pojednic.
\newblock The healthspan project: A retrospective pilot of biomarkers and biometric outcomes after a 6-month multi-modal wellness intervention.
\newblock In {\em Healthcare}, volume~12, page 676. MDPI, 2024.

\bibitem{dosovitskiy2020image}
Alexey Dosovitskiy, Lucas Beyer, Alexander Kolesnikov, Dirk Weissenborn, Xiaohua Zhai, Thomas Unterthiner, Mostafa Dehghani, Matthias Minderer, Georg Heigold, Sylvain Gelly, et~al.
\newblock An image is worth 16x16 words: Transformers for image recognition at scale.
\newblock {\em arXiv preprint arXiv:2010.11929}, 2020.

\bibitem{huang2020automatic}
Sai Huang, Rui Dai, Juanjuan Huang, Yuanyuan Yao, Yue Gao, Fan Ning, and Zhiyong Feng.
\newblock Automatic modulation classification using gated recurrent residual network.
\newblock {\em IEEE Internet of Things Journal}, 7(8):7795--7807, 2020.

\bibitem{kumar2022comprehensive}
Ashish Kumar, Rahul Saha, Mauro Conti, Gulshan Kumar, William~J Buchanan, and Tai~Hoon Kim.
\newblock A comprehensive survey of authentication methods in internet-of-things and its conjunctions.
\newblock {\em Journal of Network and Computer Applications}, 204:103414, 2022.

\bibitem{Privacy}
Linlin Li, Hui Zhu, Yandong Zheng, Fengwei Wang, Rongxing Lu, and Hui Li.
\newblock Efficient and privacy-preserving fusion based multi-biometric recognition.
\newblock In {\em GLOBECOM 2022 - 2022 IEEE Global Communications Conference}, pages 4860--4865, 2022.

\bibitem{li2023imf}
Xinhang Li, Xiangyu Zhao, Jiaxing Xu, Yong Zhang, and Chunxiao Xing.
\newblock Imf: interactive multimodal fusion model for link prediction.
\newblock In {\em Proceedings of the ACM Web Conference 2023}, pages 2572--2580, 2023.

\bibitem{lien2023challenges}
Chi-Wei Lien and Sudip Vhaduri.
\newblock Challenges and opportunities of biometric user authentication in the age of iot: A survey.
\newblock {\em ACM Computing Surveys}, 56(1):1--37, 2023.

\bibitem{liu2023time}
Chen Liu, Juntao Zhen, and Wei Shan.
\newblock Time series classification based on convolutional network with a gated linear units kernel.
\newblock {\em Engineering Applications of Artificial Intelligence}, 123:106296, 2023.

\bibitem{mamdouh2021authentication}
Moustafa Mamdouh, Ali~Ismail Awad, Ashraf~AM Khalaf, and Hesham~FA Hamed.
\newblock Authentication and identity management of ioht devices: achievements, challenges, and future directions.
\newblock {\em Computers \& Security}, 111:102491, 2021.

\bibitem{XJTU}
Dongpeng Shang, Xinman Zhang, Jiuqiang Han, and Xuebin Xu.
\newblock Multimodal-database-xjtu: An available database for biometrics recognition with its performance testing.
\newblock In {\em 2017 IEEE 3rd Information Technology and Mechatronics Engineering Conference (ITOEC)}, pages 521--526, 2017.

\bibitem{sharma2023survey}
Shreyansh Sharma, Anil Saini, and Santanu Chaudhury.
\newblock A survey on biometric cryptosystems and their applications.
\newblock {\em Computers \& Security}, page 103458, 2023.

\bibitem{talreja2020deep}
Veeru Talreja, Matthew~C Valenti, and Nasser~M Nasrabadi.
\newblock Deep hashing for secure multimodal biometrics.
\newblock {\em IEEE Transactions on Information Forensics and Security}, 16:1306--1321, 2020.

\bibitem{yang2024lutbio}
Rui Yang, Qiu-yu Zhang, Ling-tao Meng, Chun-xia Wang, and Ying-jie Hu.
\newblock {LUTBIO} multimodal biometric database, 31 July 2024.
\newblock \url{https://data.mendeley.com/datasets/jszw485f8j/2}.

\bibitem{zhang2017low}
Heng Zhang, Vishal~M Patel, and Rama Chellappa.
\newblock Low-rank and joint sparse representations for multi-modal recognition.
\newblock {\em IEEE Transactions on Image Processing}, 26(10):4741--4752, 2017.

\bibitem{Android}
Xinman Zhang, Dongxu Cheng, Pukun Jia, Yixuan Dai, and Xuebin Xu.
\newblock An efficient android-based multimodal biometric authentication system with face and voice.
\newblock {\em IEEE Access}, 8:102757--102772, 2020.

\end{thebibliography}

\end{document}